\def\year{2021}\relax
\documentclass[letterpaper]{article} 
\usepackage{aaai21}  
\usepackage{times}  
\usepackage{helvet} 
\usepackage{courier}  
\usepackage[hyphens]{url}  
\usepackage{graphicx} 
\usepackage{natbib}  
\usepackage{caption} 
\usepackage{amsmath, amssymb, amsthm, amscd, amsfonts, graphicx, booktabs, multirow}
\usepackage{subcaption}
\usepackage{makecell}
\usepackage{IEEEtrantools}
\usepackage{fge}
\usepackage{ebproof}
\usepackage{footmisc}
\usepackage{enumitem}
\usepackage{microtype}
\usepackage{latexsym}

\newcommand\ccgfslash{\textbf{/}}
\newcommand\ccgbslash{$\fgebackslash$}

\title{Generating CCG Categories}
\author {
    Yufang Liu,
    Tao Ji, 
    Yuanbin Wu, 
    Man Lan  \\
}
\affiliations {
    School of Computer Science and Technology, East China Normal University \\
    \{yfliu.antnlp, taoji.cs\}@gmail.com, \{ybwu, mlan\}@cs.ecnu.edu.cn
}

\begin{document}

\maketitle

\begin{abstract}
    Previous CCG supertaggers usually predict categories 
    using multi-class classification. 
    Despite their simplicity, 
    internal structures of categories are usually ignored.
    The rich semantics inside these structures may help us to better
    handle relations among categories and 
    bring more robustness into existing supertaggers.
    In this work, we propose to generate categories 
    rather than classify them:
    each category is decomposed into a sequence of smaller atomic tags,
    and the tagger aims to generate the correct sequence.
    We show that with this finer view on categories, 
    annotations of different categories could be shared
    and interactions with sentence contexts
    could be enhanced.
    The proposed category generator is able to achieve 
    state-of-the-art tagging ($95.5\%$ accuracy) and
    parsing ($89.8\%$ labeled F1) performances on the standard CCGBank.
    Furthermore, its performances on infrequent (even
    unseen) categories, out-of-domain texts and low resource language 
    give promising results on introducing
    generation models to the general CCG analyses.
\end{abstract}

\section{Introduction}

\begin{figure}[t]
    \centering
    \includegraphics[scale=0.45]{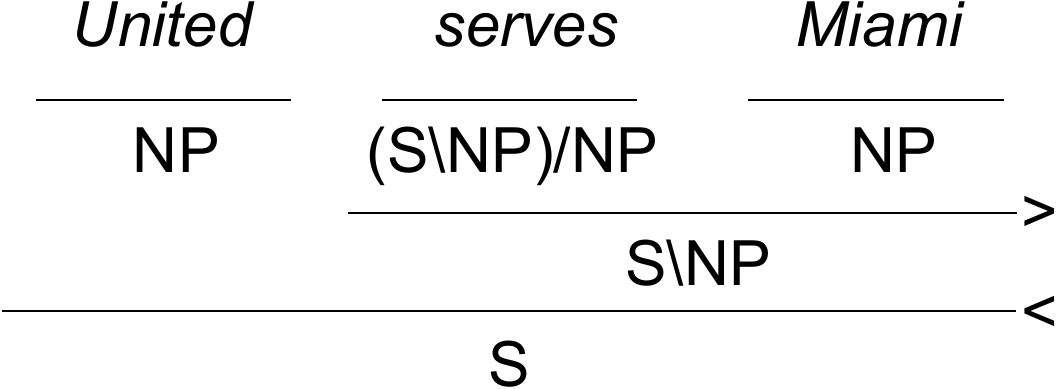}
    \caption{An example of CCG supertagging and parsing.
    ``United'' and ``Miami'' are noun phrases (NP).
    The transition verb ``serves'' has a category (supertag)
    ``(S\ccgbslash NP)\ccgfslash NP" which 
    means it first combines a right NP, then combines a left NP, and finally forms a sentence S.}
    \label{fig:ccg_intro}
\end{figure}

Supertagging is the first step of parsing natural language with
Combinatory Categorial Grammar \cite{steedman2000syntactic} 
(Figure \ref{fig:ccg_intro}). The morpho-syntax enriched
categories are of great interest not only because they can help to build 
hierarchical representations of sentences, 
but also because they provide a compact
and concise way to encode syntactic functions behind words. 
As many other data-driven NLP models, 
applications of CCG analyses are constrained with
the quality of annotations and the pre-defined category set. 
Here, for helping parsing texts in various domains,
we aim to improve the robustness of current taggers
and extend their abilities on discovering new unknown categories.


\begin{figure}[t]
    \centering
    \includegraphics[scale=0.45]{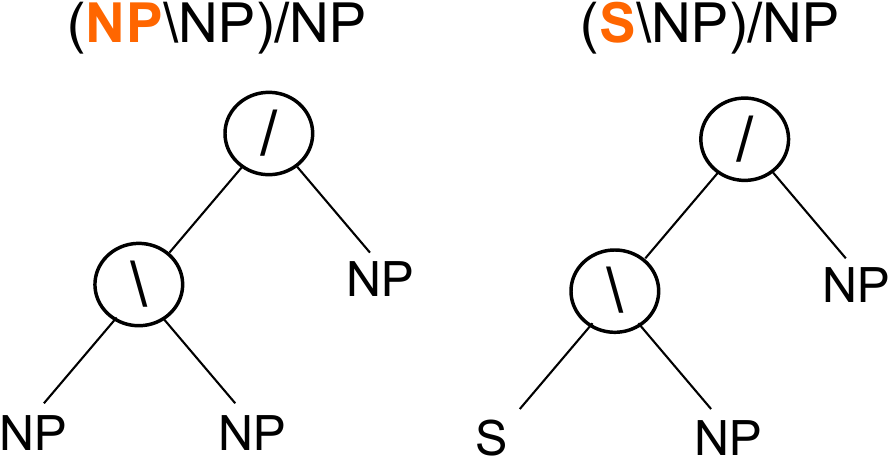}
    \caption{Two categories with similar internal structures. 
    The left represents a transitive verb,
    and the right represents a preposition attached to a noun phrase.
    }
    \label{fig:ccg_similar}
\end{figure}

The primary model for CCG supertagging is sequence labeling:
a classifier autoregressively predicts categories of sentence words.
Internal structures of categories, however, are often ignored
in this classification-based methods.
As a key feature of CCG, these structures are actually 
quite informative for handling relations among categories.
For example, the two categories 
in Figure \ref{fig:ccg_similar} are different,
but the functions they represent have the same combining strategy 
(first takes an argument from left, then from right)
and identical argument types (two NPs).
For building data-driven taggers,
this fine-grained view on categories is able to 
expose more shared information and thus helps to build a
more robust model.
For instance, 
we may rely on internal structures to improve performance of 
infrequent categories by transferring knowledge 
from more frequent categories
(which are learned more robustly).
We can also use them to induce unknown categories
by building new structures or filling new arguments,
which is impossible for existing supertaggers.

Following \citet{DBLP:conf/rep4nlp/KogkalidisMD19} work on
fine-grained type-logical category generation,
in this paper, we propose generation paradigms 
for CCG supertagging.
Instead of viewing categories as simple class labels,
we decompose them into smaller atomic tags.
Predicting a category is now equal to generate 
the corresponding atomic tag sequence.
For example, one decomposition of (NP\ccgbslash NP)\ccgfslash NP could be 
[(, NP, \ccgbslash, NP, ), \ccgfslash, NP]
which is identical to the same decomposition of 
category (S\ccgbslash NP)\ccgfslash NP
except the first NP is replaced by S.
Based on the tag sequences,
the classifier can know more
about the shared and the private learning signals of the two tags 
(e.g., by using new loss functions
based on internal structures).
It also provides a simple method to recognize new categories.

We introduce two types of category generator,
the \emph{tag-wise generator} (like NMT) which 
predicts atomic tags at each generation step
and the \emph{transition-based generator} (like parsing)
which runs a transition system to get tag sequences in
provable correct form.
We also study a spectrum of atomic tag sets 
(from the smallest tokens to the original categories,
from deterministic to non-deterministic)
to illustrate 
the potential power of category generators:
it is a flexible framework to study how categories are formed and applied.
Comparing with vanilla sentence-level sequence to sequence generation
(\citealt{DBLP:conf/rep4nlp/KogkalidisMD19}; \citealt{DBLP:conf/rep4nlp/BhargavaP20}),
the proposed generators consider hierarchical structures of categories (transition-based)
and issue multiple decoders for sentence words (faster and suitable for capturing
the property of `localized' syntax structure).


Experiments on the CCGBank show that supertagging with generation can outperform 
a strong classification baseline. 
With various decoding oracles and a simple reranker,
the tagger achieves the state-of-the-art supertagging accuracy 
(95.5\%, without using additional external resources, 96.1\% with BERT).
Furthermore, 
on low frequency and unseen categories,
the category generator is 
significantly better than the traditional category classifier.
On  out-of-domain texts (Wiki and biomedical texts) and an Italian dataset,
the category generator can also perform more robustly.




\section{Category Classifier}
\label{section.cc}

In CCG analyses, supertagging is known as \emph{almost parsing}
\citep{bangalore1999supertagging}:
most syntax ambiguities will be solved if correct 
categories (supertags) are assigned to each word in a text. 
Given a sentence $\mathbf{x}=x_1,x_2,..., x_n$, 
a supertagger predicts a tag sequence
$\mathbf{t}=t_1, t_2,...,t_n$ 
where $x_i$ is a word and $t_i$ is $x_i$'s category taking from 
a category set $\mathcal{T}$.

In this section, we introduce a typical category classification model
\citep{DBLP:conf/naacl/LewisLZ16}.
It basically first encodes sentence words into vectors, 
then performs a multi-class category classification on them.
Each word $x_i$ is mapped to a vector (also denoted by $x_i$) 
by concatenating a randomly initialized vector,
a pre-trained word embedding, and a CNN-based character embedding.
A two-layer Bi-LSTM on $\mathbf{x}$ is then applied to obtain
hidden states $h_i = [\mathop{h_i} \limits ^{\rightarrow}; \mathop{h_i} \limits ^{\leftarrow}]$,
$\mathop{h_i} \limits ^{\rightarrow} = \text{LSTM}(x_i, 
\mathop{h_{i-1}} \limits ^{\rightarrow}, \mathop{\theta} \limits ^{\rightarrow})$,
$\mathop{h_i} \limits ^{\leftarrow} = 
\text{LSTM}(x_i, \mathop{h_{i-1}} \limits ^{\leftarrow}, \mathop{\theta} \limits ^{\leftarrow})$.
After a softmax operator,
we obtain the probability of a category $p(t_i|\mathbf{x})$,
and apply the loss function $\mathcal{L} = -\sum_{i}\log p(t_i|\mathbf{x})$.

In this vanilla setting of sequence labeling, 
the relation between two tags $t'$ and $t''$ is discarded.
As a consequence, annotations of $t'$ give no
suggestion on correctly tagging $t''$.
For example, if $t', t''$
are not the true tag, they will suffer a same loss 
even one of them has more overlapping with the gold category (Figure \ref{fig:ccg_similar}). 
As we have discussed, internal structures of categories 
can make the fine-grained sharing of annotation possible. 
In the following section, 
we are going to incorporate them in the process of CCG supertagging.

\section{Category Generator}
\label{section.cg}

To explore the inner structures of categories, we first decompose them into
smaller \emph{atomic tags}\footnote{
To avoid confusion, 
we always use \emph{atomic tag} to refer to tokens in a (arbitrary) decomposition of 
original categories,
and following the CCGBank's user manual \cite{Hockenmaier05ccgbank},
we use \emph{atomic category} to refer 
categories without arguments (e.g., S, NP, N, PP,
see the manual for a full definition), which is denoted by $\mathcal{A}$.
\label{ft:atomic_notation}
}.
For example, a category (NP\ccgbslash NP)\ccgfslash NP 
(prepositions attached to noun phrases)
can be seen as a sequence,
\begin{center}
    [(, NP, \ccgbslash, NP, ), \ccgfslash, NP].
\end{center}
One advantage of such decomposition is that
now the tag has a connection with a different category (S\ccgbslash NP)/NP.
Specifically, a model can recognize that both of the categories require a left NP.
Atomic tags make these connections explicit (rather than hidden in model parameters)
and provide a way for including them in taggers.

Atomic tags can also make recognizing unknown category possible.
For categories not shown in the pre-defined label set, the classification model
can never predict them correctly.
However, by decomposing into atomic tags,
even if a category is not presented in the label set, 
it is highly possible that subsequences of the category
have been seen in the training set,
which enables the model to generate correct unknown categories.

Different from \citet{DBLP:conf/rep4nlp/KogkalidisMD19},  we propose
to deploy decoders for each individual word instead of decoding 
a single sequence for all words. Our setting may have following advantages,
first of all,  it is less sensitive to error propagation among tags
due to the decoupling of the decoding sequence. Second, tags can be parallelized
in the same sentence . 
Third, the decoder can explicitly include knowledge of the current word which fits
the idea of assigning tags to words.


Formally, we define $\mathcal{T}_a$ to be an atomic tag set of the original 
set $\mathcal{T}$ if for every category $t$ in $\mathcal{T}$, it 
can be expressed with a sequence of atomic tags in $\mathcal{T}_a$,\footnote{
An ``EOS'' is attached to every sequence as a stop sign.
}
$t=a^1,a^2,...,a^m$ where $a^j \in \mathcal{T}_a$.
For a sentence word $x_i$, our tagger's object changes to 
generate the correct sequence of atomic tags. 
We deploy two types of sequence decoders for the category generation task.
The first type follows the tag-by-tag generation paradigm.
It is simple and fast, but the generated sequences are not guaranteed to 
be well-formed.
The second type runs a transition system.
The validity of its output is guaranteed with the cost of additional 
computation steps and hard to batch.

\subsection{Tag-wise Generator}

The tag-wise generator starts an LSTM at every $x_i$.
Let $g_i^j$ be the $j$-th hidden state of the generator,
$g_i^j = \mathrm{LSTM}(g_i^{j-1}, d_i^j, \theta)$,
where 
$d_i^j = [h_i; a_i^{j-1}]$,
$h_i$ is the hidden state vector of $x_i$ from the encoder
(which keeps the generator watching $x_i$ at each step),
and $a_i^{j-1}$ represents the embedding of the output tag
from the previous generation step.
The probability of an atomic tag is defined as,
\begin{IEEEeqnarray}{c}
    \textstyle
    p(a_i^j = a| \mathbf{x}) = \mathrm{Softmax}_{a\in \mathcal{T}_a} ~ w_a^\intercal g_i^j .
    \label{eq:prob_cg}
\end{IEEEeqnarray}
The loss function is
$\mathcal{L} = -\sum_{i}\sum_{j}\log p(a_i^j|\mathbf{x})$.
Comparing with the loss of classification model,
the loss here is computed on the finer atomic tags,
which can assign credit for partially correct category predictions.

Furthermore, 
the generator is able to handle 
relations between sentence contexts and categories in a better way.
For example, 
when generating the second NP in NP\ccgfslash NP,
it might be helpful to know whether words on the left form a noun phrase.
Due to the decomposition of categories, 
each atomic tag is able to search related information from the sentence.
We apply attention layers to help such context-aware category generation.
Specifically,
at step $j$ of the category generator on word $i$,
we use hidden state $g^{j-1}_i$ to query which sentence words
are more important for predicting the next atomic tag,
$\alpha_i^{j,l} = \mathrm{Softmax}_{l\in [1, n]}( 
    w^\intercal \tanh(W_1 g^{j-1}_i + W_2 h_l) )$,
where $w, W_1, W_2$ are parameters.
A soft aggregation of all encoder vectors $h_l$ becomes a part of the
generator's input
\begin{align}
   \textstyle  d_i^j = [ h_i; ~ ~ a_i^{j-1}; ~ \sum_{l=1}^{n} \alpha_i^{j,l}h_l ].
    \label{eq:f_att}
\end{align}
In order to reduce computation costs,
we could also compute a single attention vector using $h_i$ as query
and apply it in every generation step,
$\alpha_i^{l} = \mathrm{Softmax}_{l\in [1, n]}(
    w^\intercal \tanh(W_1 h_i + W_2 h_l) )$
\begin{IEEEeqnarray}{c}
  \textstyle  d_i^j = [ h_i; ~ ~ a_i^{j-1}; ~ \sum_{l=1}^{n} \alpha_i^{l}h_l ].
    \label{eq:p_att}
\end{IEEEeqnarray}

\subsection{Transition-based Generator}
\label{section.ct}

We can also explicitly explore tree structures of categories 
during the generation.
In fact, by seeing combination operators 
(``\ccgfslash'', ``\ccgbslash'')
as non-terminals, atomic categories as terminals,
categories resemble (binarized) constituent trees. 
We can therefore adopt parsing algorithms
to obtain a well-formed categories, 
which is generally not guaranteed in tag-wise generators.
Here, we investigate an in-order transition system
\cite{liu-zhang-2017-order}, which is a variant of 
the top-down system 
\cite{dyer-etal-2016-recurrent}.

\begin{table}
    \centering
\setlength{\tabcolsep}{1em}
\begin{tabular}[t]{ll}
    axiom: & $\epsilon:0$ \\
    goal: & $\epsilon:t$ ~ ~ ($t\neq0$)
    \vspace{0.1cm} \\
    $\mathsf{gen}(a)$ & 
    {\begin{prooftree}
        \hypo{ \sigma : t }
        \infer1[~ \small $\sigma = \epsilon$ or
        top($\sigma$) is an operator]{ \sigma|a : t+1 }  
    \end{prooftree} } 
    \vspace{0.2cm} \\
    $\mathsf{op}(X)$ & 
    {\begin{prooftree}
        \hypo{\sigma|s_0 : t}
        \infer1[~ \small $s_0$ is not an operator]{\sigma|s_0|X : t+1}  
    \end{prooftree} }
    \vspace{0.2cm} \\
    $\mathsf{reduce}$ & 
    {\begin{prooftree}
        \hypo{\sigma|s_1|X|s_0 : t}
        \infer1{\sigma|X_{s_1, s_0} : t+1 }  
    \end{prooftree} }
    \vspace{0.2cm} \\
    $\mathsf{stop}$ &
    {\begin{prooftree}
        \hypo{\sigma : t}
        \infer1[~\small len($\sigma) = 1$]
        {\epsilon : t+1 }  
    \end{prooftree} } 
\end{tabular}
\caption{The transition system of generating categories. 
}
\label{tab:transition}
\end{table}

\begin{figure}
    \centering
    \small
    
     \begin{tabular}{c|l|l|l}
       \textbf{T} &\textbf{Stack}  &\textbf{Buffer} &\textbf{Action} \\
     \hline
     0 & &&$\mathsf{gen(S)}$\\
     1 & $\mathsf{S}$& $\mathsf{S}$&$\mathsf{op(\text{\ccgbslash)}}$ \\
     2 & $\mathsf{S} | \text{\ccgbslash} $  &$\mathsf{S}$& $\mathsf{gen(NP)}$ \\
     3 & $\mathsf{S} | \text{\ccgbslash} | \mathsf{NP}$ &$\mathsf{S} | \mathsf{NP} $ & $\mathsf{reduce}$ \\ 
     4 & $\mathsf{S\text{\ccgbslash}NP}$ &$\mathsf{S} | \mathsf{NP} $& $\mathsf{op(\text{\ccgfslash)}}$\\
     5 & $\mathsf{S\text{\ccgbslash}NP | \text{\ccgfslash}} $ &$\mathsf{S} | \mathsf{NP} $&$\mathsf{gen(NP)}$ \\
     6 & $\mathsf{S\text{\ccgbslash}NP | \text{\ccgfslash} | \mathsf{NP}}$ &$\mathsf{S} | \mathsf{NP} | \mathsf{NP}$&$\mathsf{reduce}$\\
     7 & $(\mathsf{S\text{\ccgbslash}NP) \text{\ccgfslash} NP}$ &$\mathsf{S} | \mathsf{NP} | \mathsf{NP}$& $\mathsf{stop}$\\
    \hline
  \end{tabular}
    
    \caption{An example of category $(\mathsf{S\text{\ccgbslash}NP) \text{\ccgfslash} NP}$ for transition-based generator.}
    \label{fig:transition_example}
\end{figure}

Table \ref{tab:transition} illustrates the deduction 
rules of the transition-based generator.
Each transition state contains a stack $\sigma$ and the current timestep $t$.
$\mathsf{gen}(a)$ generates an atomic category $a \in \mathcal{A}$\footref{ft:atomic_notation} and pushes $a$ to the stack $\sigma$.
$\mathsf{op}(X)$ generates a combination operator 
$X \in \{\text{\ccgfslash}, \text{\ccgbslash}\}$ and push $X$ to $\sigma$.
$\mathsf{reduce}$ combines the top three elements of $\sigma$ 
and concatenates them to the output $t$.
$\mathsf{stop}$ is the stopping rule.
An example of transition is shown in Figure \ref{fig:transition_example}.

At each step of the generation,
a classifier predicts which action to perform.
Following \citet{dyer-etal-2016-recurrent}, 
we use a stack-LSTM to encode stack states.
The detailed configuration is in the supplementary
due to the lack of space.

\paragraph{Discussions} The transition-based generator produces categories with provably
correct form, which is not guaranteed in the tag-wise generator.
On the other side, the tag-wise generator is easier to
batch and much faster.
Empirically, we find that the problem of illegal categories is not severe 
in the tag-wise generation: all 1-best outputs of the generator are legal
and only 0.05\% of 4-best outputs are wrong.
In fact, like recent practice of sequence-style parsing 
\cite{zhang-etal-2017-dependency-parsing,fernandez-gonzalez-gomez-rodriguez-2019-left,shen-etal-2018-straight},
it is possible to drop structure constraints 
with a well-learned sequence decoder.
Categories are usually short (average length is $4$) 
and their number is also limited ($10^3$).
All these factors increase the chance of obtaining well-formed
categories directly from the tag-wise generator.
We thus focus on this simpler implementation.

We also note that it's straightforward to apply 
advanced encoder structures(in fact, we apply
BERT\cite{DBLP:conf/naacl/DevlinCLT19} in our experiments).
However, we would like to think the main contribution here is
to study CCG Supertagging from a new perspective,
rather than a new generation model.

\subsection{Decoding Oracles}
\label{section: decoding oracles}

One key point in category generators
is how to define the atomic tag set $\mathcal{T}_a$
which determines the learning targets (\emph{oracles}) of the decoder.
For the transition-based generator, 
$\mathcal{T}_a$ is simply the transition action set.
In the following, we are going to show different settings of 
$\mathcal{T}_a$ for the tag-wise generator.
Following the semantics of CCG, we have a natural choice of 
$\mathcal{T}_a$,
\begin{align}
    \textstyle
    \mathcal{T}_a = \mathcal{A}\cup \{(, ), \text{\ccgbslash}, \text{\ccgfslash} \}, 
    \tag{\textbf{AC}}\label{eq:oracle_ac}
\end{align}
where $\mathcal{A}$ contains atomic categories\footref{ft:atomic_notation} of the grammar.
 
It's easy to see that each category $t \in \mathcal{T}$ corresponds to a unique 
atomic tag sequence from \ref{eq:oracle_ac},
which forms a \emph{deterministic oracle} for the category generator.

We can enrich \ref{eq:oracle_ac}, for example,
with some parentheses expressions 
(e.g., ``NP\ccgbslash NP'' in category 
``(NP\ccgbslash NP)\ccgfslash NP'') in the original category
(which may help to handle some common local syntactic functions),
\begin{align}
    \textstyle
    \mathcal{T}_a = \textbf{AC}\cup \mathcal{P}_k,
    \tag{\textbf{PA}}\label{eq:oracle_pa}
\end{align}
where $\mathcal{P} = \{\tau|(\tau) \text{ is a substring of a } 
t\in \mathcal{T}\}$,
$\mathcal{P}_k$ is the subset of $\mathcal{P}$ with top-$k$ frequent items.

Furthermore, we could also either completely ignore 
the semantics of categories by adding their $n$-grams
or completely accept them by adding all items in $\mathcal{T}$,
\begin{align}
    \textstyle
    \mathcal{T}_a &= \textbf{AC}\cup \mathcal{N}_k^n,
    \tag{\textbf{NG}}\label{eq:oracle_ng} \\
    \mathcal{T}_a &= \textbf{AC}\cup \mathcal{T},
    \tag{\textbf{OR}}\label{eq:oracle_or}
\end{align}
where $\mathcal{N}^n = \{\tau|\tau \text{ is a } n\text{-gram of a }  
t\in \mathcal{T}\}$, 
$\mathcal{N}^n_k$ is the subset of $\mathcal{N}$ with top-$k$ frequent $n$-grams.

Unlike \ref{eq:oracle_ac}, when $\mathcal{T}_a$ is set to
\ref{eq:oracle_pa}, \ref{eq:oracle_ng} and \ref{eq:oracle_or},
a category $t$ may have more than one correct sequences.
For example, with \ref{eq:oracle_pa}, the tag (NP\ccgfslash NP)\ccgbslash NP may have
two gold standard atomic tag sequences,
[(, NP\ccgfslash NP, ), \ccgbslash, NP]
and [(, NP, \ccgfslash, NP, ), \ccgbslash, NP].

We can still pick a deterministic oracle by applying some heuristic
rules. Here, the deterministic oracles always perform the longest forward matching
(i.e., with a prefix $a^1, a^2,\dots,a^j$, $a^{j+1}$ is set to 
a feasible atomic tag with the longest length).\footnote{
We assume there is only one tag with the longest length.
}
On the other hand, we also investigate \emph{non-deterministic oracles}
for training the tag-wise generator.
Instead of using a fixed oracle during the entire training process, 
we select oracles randomly for each category,
and all oracles will participate in the learning of the supertagger.

\section{Re-ranker}
\label{section.reranker}

To combine a category generator
and the category classifier,
we further introduce a simple re-ranker.
First, using beam search, we can obtain $k$-best categories 
from the category generator.
For each category $t=a^1,a^2,\cdots,a^m$,
we assign it a confidence score using probabilities of tags
(Equation \ref{eq:prob_cg}),
$u_t = \frac{1}{m^\nu}\sum_{j=1}^m{\log{p(a^j|\mathbf{x})}}$
where $\nu \leq 1$ is a hyperparameter using to penalize long 
tag sequences.

Next, we use the category classifier to obtain
category $t$'s probability $\log p(t|\mathbf{x})$ 
as its confidence score $v_t$.
The final score of $t$ is defined as the weighted sum of the two scores
$\lambda u_t + (1 - \lambda) v_t$. 
The category with the highest score is taken as the final output.
We set $\nu=0.15, \lambda=0.9$ by selecting them on the development data.

\section{Experiments}

\begin{table}
  \centering
  \resizebox{\linewidth}{!}{
  \begin{tabular}{lllcc}
    \toprule
    \textbf{Model} & \textbf{Dev} & \textbf{Test} & \textbf{Size} & \textbf{Speed}\\
    \midrule
    C\&C  & 91.50  & 92.02 & - & \\
    \citet{DBLP:conf/naacl/LewisLZ16} & 94.10 & 94.30 & 48.88 & -\\
    \quad \quad \quad +tri-training  & 94.90 & 94.70 & -& -\\
    \citet{DBLP:conf/naacl/VaswaniBSM16} & 94.24 & 94.50 & - & -\\
    \citet{DBLP:conf/aaai/WuZZ17} & 94.50 & 94.71 & 99.16 & -\\
    \citet{DBLP:conf/nlpcc/WuZZ17} & 94.72 & 95.08 & 189.37 &-\\
    \hline
    CC  & 94.89 & 95.21&77.11& 466\\
    CG  & 95.10 & 95.28&79.94& 199\\
    CGNG2 & 95.26$^*$ & 95.44$^*$&80.02& 199\\
    CT &94.06&94.09&77.97& 21\\
    rerank &\textbf{95.27}$^*$&\textbf{95.48}$^*$&-&199\\
    \midrule
    \midrule
    \textbf{Pre-training}&&&&\\
    \midrule
    \citet{DBLP:conf/emnlp/ClarkLML18} &-&96.10&-&-\\
    \citet{DBLP:conf/rep4nlp/BhargavaP20}&\textbf{96.27}&96.00&-&-\\
    \midrule
    BERT+CC& 96.01& 95.93&78.62&231\\
    BERT+CG& 96.13& 95.97&81.35&131\\
    BERT+CGNG2& 96.18& 95.99&81.53&131\\
    BERT+CT&95.28&94.91&79.48&17\\
    BERT+rerank&96.24$^*$&\textbf{96.05}$^*$&-&131\\
    \bottomrule
  \end{tabular}
  }
  \caption{Comparing with existing supertaggers.
    Model sizes are the number of parameters (MB).
    Speeds are in sentence per second.
    We use BERT-base without fine-tuning.
    All results of our models are
     averaged over 3 runs.
     * indicates significantly better.
    }
  \label{table:all_results}
\end{table}

\paragraph{Datasets and Criteria}
We conduct experiments mainly on CCGBank \citep{DBLP:journals/coling/HockenmaierS07}.
We follow the standard splits of CCGBank using section 02-21 for training set,
section 00 for development set, and section 23 for test set.
There are $1285$ different categories in training set, 
following the previous taggers, we only choose $425$ of them
which appear no less than $10$ times in the training set, 
and assign \texttt{UNK} to the remaining tags.

For out-of-domain evaluation,  we use the Wikipedia corpus 
\cite{Clark2009LargeScaleSP} and the Bioinfer corpus 
\cite{DBLP:journals/jbi/RimellC09}.\footnote{They include 1000 Wikipedia sentences and 1000 biomedical (GENIA) sentences 
with noun compounds analysed. 
} 
We also test our models on the news corpus of the Italian CCGBank\cite{bos2009converting},
We use the token-POS-category tuples file from the Italian news corpus.\footnote{
We use  period to spilt the dataset and get $740$ sentences as train/dev/test(8:1:1).
Dataset can be download from \url{http://www.di.unito.it/~tutreeb/CCG-TUT/.}}

The main criterion for evaluation is tag accuracy.
To measure statistical significance,
we employ t-test \citep{DBLP:conf/acl/ReichartDBS18}
with $p < 0.05$.\footnote{\url{https://github.com/rtmdrr/testSignificanceNLP}}
The settings of network hyperparameters are in the supplementary.
We compare several models,
\begin{itemize}[leftmargin=*]
    \item CC, the category classifier in 
        Section \ref{section.cc}.
    \item CG, the tag-wise generator with deterministic oracle 
        \ref{eq:oracle_ac}.
    \item CGNG2, the tag-wise generator with deterministic 
        \ref{eq:oracle_ng} ($k=10,n=2$).
    \item CT, the transition-based system in Section \ref{section.ct}.
    \item rerank, combining CG and CGNG2 with the ranker (beam size is $4$).
\end{itemize}

\subsection{Main Results}
\label{section.main_results}


Table \ref{table:all_results} lists overall performances on CCGBank.
C\&C is a non-neural-network-based CCG parser,
\cite{DBLP:conf/naacl/LewisLZ16} is a LSTM-based supertagger 
similar to our CC model (with less parameters).
It also uses tri-training-based semi-supervised learning 
\cite{DBLP:conf/acl/WeissACP15}.
Shortcut LSTM \citep{DBLP:conf/nlpcc/WuZZ17}
performs best in previous works, which uses the shortcut block as a basic architecture 
for constructing deep stacked models. 
Their final model uses 9-layer stacked shortcut block as encoder.
And a contemporaneous work \citep{DBLP:conf/rep4nlp/BhargavaP20}
which use a single decoder for the whole sentence.
From the results, we find that,
\begin{itemize}[leftmargin=*]
    \setlength\itemsep{0em}
    \item Our implementation of the category classifier (CC)
        outperforms the best previous system (Shortcut LSTM)
        with much less parameters.
    \item With the same encoder and a small increase of model size,
        tag-wise generators could
        bring further performance gains (CG and CGNG2).
        However, our current transition-based generator underperforms
        the classification model.
        Regarding the implementation of transition systems, 
        we adopt the standard stack-LSTM which doesn't fully explore 
        the features of transition structures. 
        It is possible that further feature engineering and advanced encoders 
        will improve the performances.
        Finally, the reranker can reach a new state-of-the-art
        in supertaggers using no external data.
    \item Regarding tagging speeds, 
        since tag-wise generators need additional decoding steps
        on sentence words,
        they speeds are roughly two-fifths of the classification model. 
        The transition-based generator 
        is much slower since it needs to build features from the
        stack, the current output and history actions 
        using LSTMs at every decoding step.
    \item All of our models obtain an appreciable increase 
        in performance with the help of BERT \cite{DBLP:conf/naacl/DevlinCLT19}.
        The results of NGCG2 (with rerank) are comparable 
        to the results of cross view training
        \cite{DBLP:conf/emnlp/ClarkLML18} which uses 
        unsupervised data and annotations from other tasks
        and the contemporaneous work \citep{DBLP:conf/rep4nlp/BhargavaP20} 
        which shares the same idea of generating categories.
    \item We also test our models on the Italian CCGBank, it shows there is no
        significant difference between the results of CC 
        and CG models.
        And our CGNG2 model performs best(64.10\%). It proves that our tag-wise
        generators can still perform well with few data. 
        Detailed results are in the supplementary. 
\end{itemize}

\begin{table}
  \centering
  \small
  \begin{tabular}{c|lcc}
    \toprule
     & \textbf{oracles} & \textbf{Dev} & \textbf{Test} \\
    \midrule
    CC & - & 94.89 & 95.21 \\
    \midrule
    \multirow{5}{*}{D }
    & \ref{eq:oracle_ac}  & 95.10          & 95.28          \\
    & \ref{eq:oracle_pa}     & 95.12          & 95.31          \\
    & \ref{eq:oracle_ng}($n=2$)    & \textbf{95.26} & 95.44          \\
    & \ref{eq:oracle_ng}($n=3$)    & 95.13          & 95.37          \\
    & \ref{eq:oracle_ng}($n=4$)    & 95.25          & \textbf{95.48} \\ 
    \hline
    \multirow{5}{*}{ND}
    & \ref{eq:oracle_pa}    & 95.08          & 95.43          \\
    & \ref{eq:oracle_or} & 95.00          & 95.24          \\
    & \ref{eq:oracle_ng}($n=2$)    & 95.07          & 95.39          \\
    & \ref{eq:oracle_ng}($n=3$)    & 95.23          & 95.38          \\
    & \ref{eq:oracle_ng}($n=4$)    & 95.20          & 95.35          \\
    \bottomrule
  \end{tabular}
  \caption{Results of tag-wise generators combined with various oracles.
    D and ND represent deterministic and non-deterministic oracles.
    For PA and NG, we set $k=10$.
    Except AC and OR, the improvements
    are significant (with $p < 0.05$).
    }
  \label{table:various oracles}
\end{table}


Next, we show performances of the tag-wise generator with
different oracles (Table \ref{table:various oracles}).
In general, comparing with the category classifier,
the sequence oracles could 
effectively boost tagging accuracies.
The following are some observations.
\begin{itemize}[leftmargin=*]
    \setlength\itemsep{0em}
    \item It is interesting to see that $n$-gram oracles \ref{eq:oracle_ng}
    perform better (on Dev) than other oracles both on
    deterministic and non-deterministic settings.
    We guess that, besides 
    existing atomic categories in \ref{eq:oracle_ac} 
    (and their simple combinations in \ref{eq:oracle_pa}),
    which have clear definitions from linguistic prior,
    there still exist some other latent linguistic structures 
    which might help CCG analyses.
    How to uncover them is our important future work.
    \item Except \ref{eq:oracle_ng} ($n=3$),
    the non-deterministic oracle
    is not able to get better accuracies than deterministic oracles.
    One reason might be that simply using 
    random learning targets may make the generator harder to learn,
    thus more advanced fusion strategies are desired.
    \item We have tested oracle \ref{eq:oracle_ac} and \ref{eq:oracle_ng}
    with larger $k$ (i.e., including more items).
    The results are similar to those in Table \ref{table:various oracles}, 
    which may suggest that  
    the oracles are not quite sensitive to items' frequencies
    when choosing properly.
\end{itemize}

Third, we show the effectiveness of attention layers.
Constrained by our hardware platform,
instead of using the default setting, we evaluate a smaller model
(the batch size becomes $128$, 
the dimensions of the encoder and the decoder LSTM are decreased to $300$ and $200$).
The results (Table \ref{tab:ablation_study})
show that, though attention layers require more computation resources, 
they can help to achieve significantly better tagging accuracies
than the vanilla category generator.
The two different attention settings (Equation \ref{eq:f_att}, \ref{eq:p_att})
performs nearly the same (thus we may prefer the faster one 
(Equation \ref{eq:p_att})).

Finally, we test the  percentages of illegal
categories generated from category generators, the results show that 
all 1-best outputs of CC and NGCG2 are legal and only $0.05\%$, $0.04\%$ 
of 4-best outputs are wrong.
It suggests that it is not hard for tag-wise generators 
to build well-formed categories given our moderate capacity
decoding structures.

\begin{table}
  \centering
  \small
  \begin{tabular}{p{4.0cm}  p {1.6cm}}
    \toprule
    \textbf{Model}            & \textbf{Acc}   \\
    \midrule
    CG$^\triangle$  &94.30 \\
    \quad + attention(Equation \ref{eq:p_att}) &\quad +0.63\\
    \quad + attention(Equation \ref{eq:f_att}) &\quad +0.65\\
    CC               & 94.89  \\
    \quad  w/o cnn    & \quad -0.19 \\
    \quad w/o dropout      & \quad -1.49 \\
    CG            & 95.10 \\
    \quad  w/o cnn      & \quad -0.36 \\
    \quad w/o dropout  &  \quad -1.30\\
    \quad w/o tag embedding & \quad -0.59\\
    \bottomrule
  \end{tabular}
  \caption{Ablation studies.
  The last row means the model without atomic tag embedding 
  in the last decoding step. 
  $\triangle$ denotes a smaller model for running attention.
  }
  \label{tab:ablation_study}
\end{table}

\subsection{Robustness}


\begin{table}[t]
  \small
   \centering
   \begin{tabular}{llll}
     \toprule
       & \textbf{10} $\sim$ \textbf{100}            & \textbf{100} $\sim$ \textbf{400} & \textbf{400} $\sim$ \textbf{2000}  \\
     \midrule
     CC      &  60.41& 77.06 &     86.77     \\
     CT & 41.86 &63.81&79.73\\
     CG        & 62.44   & 77.51 &   87.58$^*$   \\
     CGNG2    &   \textbf{65.83}$^*$  & 78.95$^*$  &       87.93$^*$    \\
     rerank &64.25$^*$&\textbf{79.84}$^*$&\textbf{88.16}$^*$\\
     \% in test & 40.46\%&17.70\%& 11.49\%\\
     \bottomrule
   \end{tabular}
   \caption{Accuracy of infrequent categories on the test set. 
   We group categories with their frequency in the training set.
   The last row shows the proportion of categories
   in the test set.
   * indicates significantly better than model CC.}
   \label{table: few_shot_results}
 \end{table}

\begin{figure}[t]
  \centering
  \includegraphics[scale=0.17]{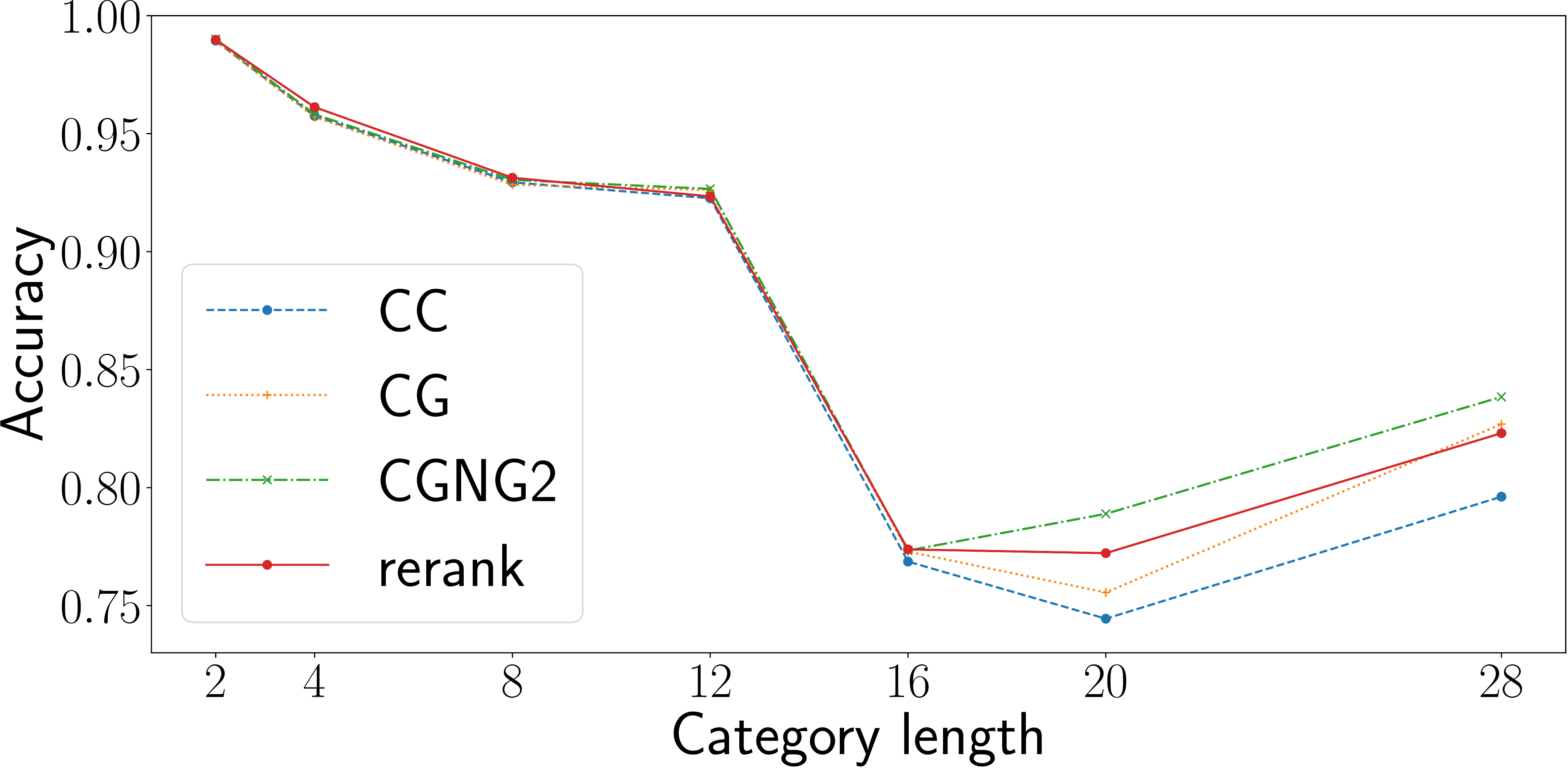}
  \caption{Accuracy on categories of different lengths. 
  }
  \label{fig:category_len}
\end{figure}

By inspecting the CCGBank training set,
we see that there are about two-thirds of categories 
which appear less than $10$ times ($1-425/1285$),
and more than half of the remaining categories appear less than $100$ times 
($253/425$). 
We now show
how the category generator performs on these infrequent (or even unknown)
categories, which can be a sign for model robustness.

First, from Table \ref{table: few_shot_results},
category generators exhibit significantly better tagging accuracies
on infrequent categories, and as the annotation becomes less,
the gap between generators and the classifier becomes larger.
Similarly, 
we compare systems with respect to the length of a category
(empirically, a longer length implies a small frequency).
Figure \ref{fig:category_len} shows that for categories
with length less than $16$, the models perform almost identical,
but for longer categories,
the category generator give more robust tagging results
(especially for CGNG2 which has fewer generation steps than CG).


\begin{table}[t]
    \small
  \centering
  \begin{tabular}{lcccc}
  \toprule
       &\textbf{p@1} &\textbf{p@2} &\textbf{p@4} &\textbf{p@8}\\
     \midrule
      CG  &11.54 &17.31 &20.19&25.00\\
    \quad w/o feature & 20.19 & 32.69& 35.58 &43.27\\
     CGNG2 &8.65  & 14.42& 15.38&22.12\\
     \quad w/o feature & 21.15 & 29.81&31.73&39.42\\
     CT &0.96&4.81&5.77&7.69\\
     \quad w/o feature &6.73&14.42&25.96&31.73\\
    \bottomrule
  \end{tabular}
  \caption{The results on unknown categories.
  ``p@k'' measures whether the correct category appears
  in the top-k outputs of category generators.
  ``w/o feature'' means when comparing categories,
  we ignore their features (e.g., S[dcl] equals S).}
  \label{table:zero_shot_results}
\end{table}

\begin{table}[t]
  \small
  \centering
  \begin{tabular}{cc}
    \toprule
    \textbf{Category}  & \textbf{Prediction} \\
    \midrule
    S[wq]\ccgfslash N         & S[wq]\ccgfslash(S[q]\ccgfslash NP)       \\
    (NP\ccgfslash NP)\ccgfslash N   & NP[nb]\ccgfslash N                      \\
    conj\ccgfslash PP                            & conj                          \\
    (((S[pt]\ccgbslash NP)\ccgfslash PP)\ccgfslash PP)\ccgfslash NP &
    ((S[pt]\ccgbslash NP)\ccgfslash PP)\ccgfslash NP \\
    N\ccgfslash S[qem]      & N\ccgfslash S[dcl]    \\
    S[wq]\ccgfslash S[dcl]  & S[wq]\ccgfslash S[q]                    \\
    (N\ccgbslash N)\ccgfslash(N\ccgbslash N)  & conj    \\
    \bottomrule
  \end{tabular}
  \caption{Some examples of prediction on categories not in the training label set.}
  \label{table: zero_shot_examples}
\end{table}

Next, we show performances of category generators on unknown categories.
Recall that we only use $425$ of $1285$ categories in the training set
(the remaining categories are tagged with \texttt{UNK},
and we still test all categories on test set)\footnote{
  We actually do experiments on models training on the whole tag set,
  the results are almost the same. Considering there are too few (specifically, 22)
  unseen categories which are not shown in the $1285$ tag set  
  to test performances. Thus we finally choose to train on the $425$ categories.
}. 
In order to avoid models considering \texttt{UNK} tag as a true tag, 
for \texttt{UNK} tags in the training set, 
we exclude their loss during the training
(they are still fed into encoders 
in order to not break the input sentence).
On the test set, there are $104$ words with categories not 
included in the $425$ training tags,
and we show the results on these tags in 
Table \ref{table:zero_shot_results}.
We can observe that, given the top-k candidates, 
the unseen tags can have a chance to be included,
thus the generator might be a reasonable method
to deal with unseen categories.
We also find that CGNG2 now has lower performances comparing
with CG.
One reason might be that when generating unseen categories,
due to the lack of prior knowledge, the semantic of 
original atomic categories (established by linguists) 
are more important than the implicit (raw) information 
hidden in n-gram tags.


Some failed examples of generating unknown category are shown in 
Table \ref{table: zero_shot_examples}.
In the first and second lines,
CG gives partially correct results.
In the third and fourth line, an argument (PP) is missing.
In the fifth and sixth lines, 
the prediction is mostly right except for wrong features of S
(a declarative sentence is predicted as a yes-no question,
since we have no special treatment on features of categories, 
it could be further improved).
CG is completely wrong in the last row.

\begin{figure}[t]
  \centering
  \includegraphics[scale=0.18]{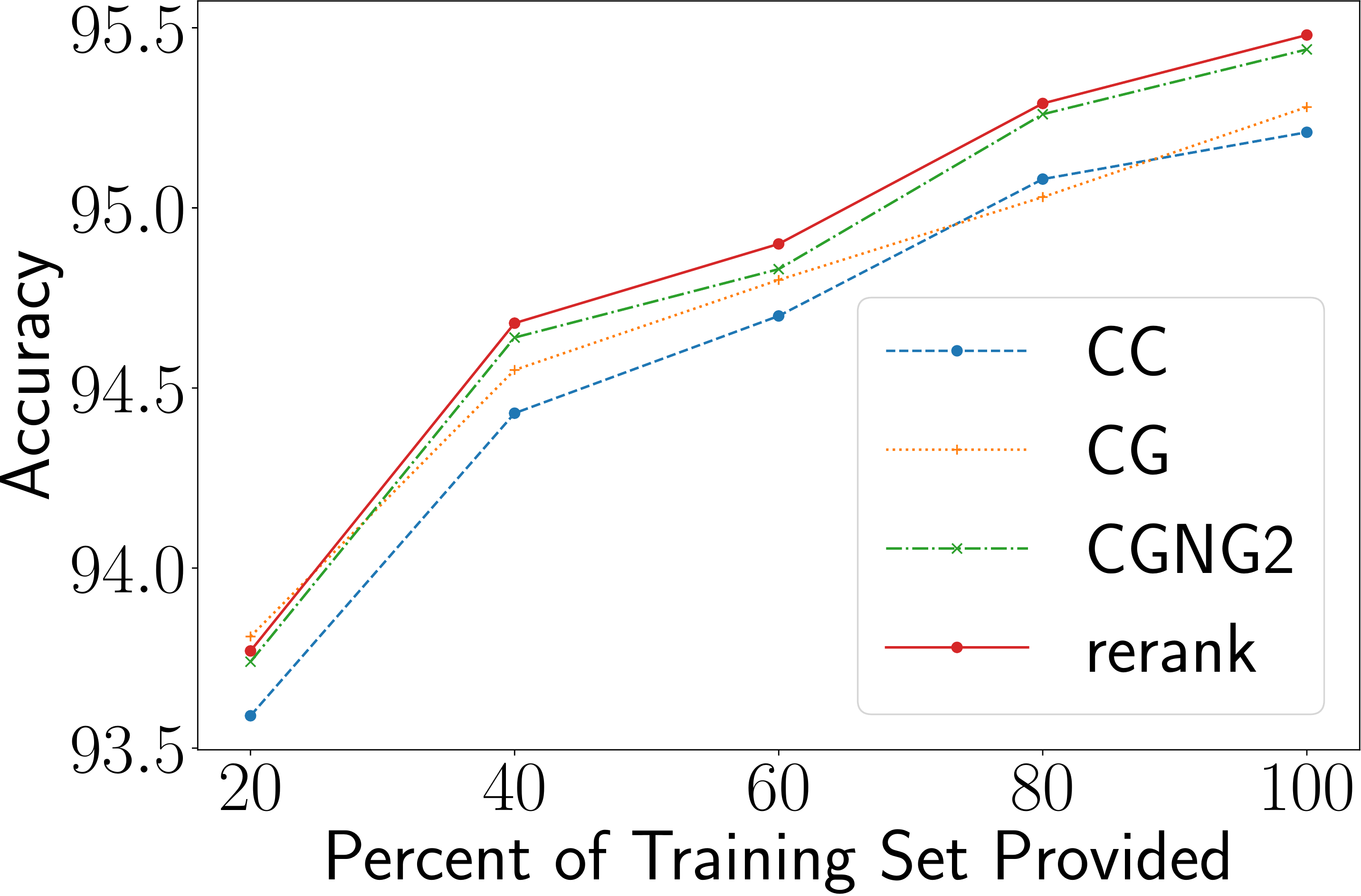}
  \caption{Tagging accuracy on test set with different training set sizes.
  }
  \label{fig:data_dec}
\end{figure}

We also show overall performances when we reduce
the size of the training set in Figure \ref{fig:data_dec}
(which may not increase the number of unknown tags, 
but provide an approximate setting).
The generation model consistently outperforms
the classification model with limited training data.

Then, we show the tagging results on out-of-domain
data (Table \ref{table: out_of_domain_results})
using models trained on the CCGBank.
We find that CG performs significantly better than the baseline CC model.
Therefore, the robustness of category generator can also extend
to texts in different domains.


\begin{table}[t]
    \small
  \centering
  \begin{tabular}{lll}
  \toprule
  \textbf{Method} &\textbf{Bioinfer} &\textbf{Wiki} \\
     \midrule
      CC  &80.68 & 92.05\\
      CG & 80.68  & 92.24$^*$\\
      CGNG2 & 80.99$^*$& 92.34$^*$\\
      rerank &\textbf{81.05}$^*$&\textbf{92.42}$^*$\\
    \bottomrule
  \end{tabular}
  \caption{Results on out-of-domain data sets. 
  $^*$ denotes the difference
  between one model and the CC model is significant.}
  \label{table: out_of_domain_results}
\end{table}

\subsection{Parsing Results}

\begin{table}[t]
  \centering
  \resizebox{\linewidth}{!}{
  \begin{tabular}{lcccc}
  \toprule
  \textbf{Model} &\textbf{F1} &\textbf{UF1}&\textbf{F1$^{\dag}$}&\textbf{UF1$^{\dag}$} \\
    \midrule
    C\&C & 85.45 &91.65  &-&-\\
    \citet{DBLP:conf/emnlp/LewisS14} &83.37&  -&-&-\\
    \citet{DBLP:conf/acl/XuAC15} &87.04 &-&-&-\\
    \citet{DBLP:conf/naacl/LewisLZ16} &87.80&-&-&-\\
    \citet{DBLP:conf/naacl/VaswaniBSM16}&88.32&-&-&-\\
    \citet{DBLP:conf/acl/YoshikawaNM17}&88.80&94.00&-&-\\
     \midrule
      CC &89.52&94.05&90.69&94.71\\
      CG  & 89.68& 94.14&90.77&94.76\\
      CGNG2 &89.76& 94.22&90.82&94.79\\
      CT&88.37&93.36&89.70&94.17\\
     rerank& \textbf{89.80}&\textbf{94.22}&\textbf{90.87}&\textbf{94.83}\\
     \midrule
      \textbf{Gold pos tag} &&&&\\
     \citet{DBLP:conf/rep4nlp/BhargavaP20}  &90.2&-&90.90&-\\
     rerank &\textbf{90.24}&94.52&\textbf{91.15}&95.01\\
    \bottomrule
  \end{tabular}
  }
  \caption{Parsing results on test set. $^{\dag}$ means using BERT.
   All results of our models are averaged over 3 runs. }
  \label{table:parsing_results}
\end{table}

To show the CCG parsing performances(Table \ref{table:parsing_results}),
we feed outputs of our supertaggers into 
the C\&C parser \cite{DBLP:journals/coling/ClarkC07}.
We compare our models with 
the C\&C parser with a RNN supertagger
\cite{DBLP:conf/acl/XuAC15},
the A* parser with a feed-forward neural network supertagger
\cite{DBLP:journals/tacl/LewisS14},
the A* parser with a LSTM supertagger 
\cite{DBLP:conf/naacl/LewisLZ16},
the A* parser with a language model enhanced biLSTM supertagger
\cite{DBLP:conf/naacl/VaswaniBSM16},
the A* CCG parser with a factorized biLSTM supertagger
\cite{DBLP:conf/acl/YoshikawaNM17},
and C\&C parser with a category generator for the whole sentence
\cite{DBLP:conf/rep4nlp/BhargavaP20}.

In general, the parsing performances are consistent with 
supertagging results in Table \ref{table:all_results}:
the rerank model achieves the best labeled and unlabeled
parsing results.
We also see that, even tag-wise generators may output 
illegal outputs, their parsing performances are better
than the transition-based generator.
An explanation, in addition to better supertagging
results, is that CCG parsers are able to 
utilize k-best supertagger sequences (which further reduce 
the influence of one single illegal category) 
and ignore ill-formed categories easily (as the combination rules 
are always non-applicable to them).

\section{Related Work}

Traditionally, CCG supertagging is seen as a sequential labelling task.
\citet{DBLP:journals/coling/ClarkC07}
propose C\&C tagger which uses
a log-linear model to build the supertagger.
Recent works have applied neural networks to supertagging (\citealt{DBLP:conf/acl/XuAC15}; 
\citealt{DBLP:conf/naacl/VaswaniBSM16}; \citealt{DBLP:conf/nlpcc/WuZZ17}). 
These works perform a multi-class classification on 
pre-defined category sets and they can't capture the inside connections
between categories because categories are independent of each other.
\citet{DBLP:conf/emnlp/ClarkLML18} propose Cross-View Training 
to learn the representations of sentences, which effectively leverages 
predictions on unlabeled data and achieves the best result.
However, their model needs a large amount of unlabeled data. 
\citet{DBLP:conf/naacl/VaswaniBSM16} also want to model the interactions between supertags, 
but unlike our methods they use a language model to capture these connections.
The difference is that we no longer treat every category as 
a label but a sequence of atomic tags.

The work closest to ours is \citet{DBLP:conf/rep4nlp/BhargavaP20}. We share the same idea of 
generating categories but there are still some key differences. They decode
a single sequence for all words while we deploy decoders for each individual words which 
may solve some problems(see Section~\ref{section.cg}).
Besides, \citet{DBLP:journals/corr/abs-2012-01285} also investigate the internal structure
of CCG supertag. They treat each category as a single tree (just like our transition system)
and use TreeRNNs for tree-structured category prediction.

Seq2Seq model has been used in many NLP tasks, such as  machine translation
(\citealt{DBLP:conf/nips/SutskeverVL14}; \citealt{DBLP:journals/corr/BahdanauCB14}),
text summarization (\citealt{DBLP:conf/conll/NallapatiZSGX16};
\citealt{DBLP:conf/acl/SeeLM17}), and especially on syntax parsing.
More related, \citet{DBLP:conf/nips/VinyalsKKPSH15} and \citet{DBLP:conf/aaai/MaLTZS17}
use Seq2Seq model to generate constituency grammar,
and \citet{DBLP:conf/coling/LiCHZ18}, 
\citet{DBLP:conf/emnlp/ZhangLLZC17} use Seq2Seq model to generate dependency grammar.
Inspired by their works, we apply Seq2Seq model to generate CCG supertags.
But the difference  is our generation is token level while theirs are sentence level.
By splitting categories into smaller units, we decrease the size of label set. And the results show
our category generating model performs well.



Techniques for classifying the unseen label  have been investigated in many tasks, 
such as  computer vision (\citealt{DBLP:journals/pami/TorralbaMF07};
\citealt{DBLP:conf/cvpr/BartU05};
\citealt{DBLP:journals/pami/LampertNH14})
and transfer learning (\citealt{DBLP:conf/eccv/YuA10};
\citealt{DBLP:conf/cvpr/RohrbachSS11}).
It would be an important future work to 
introduce advanced algorithms for dealing with these unknown categories.


\section{Conclusion}
We proposed a category generator to improve supertagging performance.
It provides a new way to capture relations among different categories
and recognizing unseen categories.
We studied a Seq2Seq-based model,
as well as a set of learning targets for the generator.
Experiments on CCGBank, out-of-domain datasets and an Italian dataset 
show the effectiveness of our model. 
Future work will explore improving the accuracy of
non-deterministic oracle and different rerankers.
We will also study how to further improve tagging infrequent categories.

\section{Acknowledgments}
We would like to thank all reviewers for their helpful comments and suggestions. 
The corresponding author is Yuanbin Wu. 
This research is (partially) supported by NSFC (62076097), STCSM (18ZR1411500, 
19511120200), and the Foundation of State Key Laboratory of Cognitive Intelligence, 
iFLYTEK(COGOS-20190003).

\bibliography{aaai}

\end{document}


\maketitle

\section{Transition-based Generator}
\label{section.ct}

As we described, we make a list of rules to guarantee the generated categories are legal.
Except from the main rules showed before, there are extra two  rules for punctuation characters
\footnote{There are some punctuation characters in CCGBank, such as period, comma, semi-colon and colon.}:
when the stack is not empty, the punctuation characters can not be generated;
if punctuation characters have been generated, the operation can only be $\mathsf{stop}$.
Obviously, under the control of all the  rules, the model can only generate
sequences of alternating terminal symbols and non-terminal symbols, which can be
transformed to a legal categories easily.

\begin{figure}[t]
    \centering
    \begin{minipage}[t]{0.45\textwidth}
    \centering
    \includegraphics[width=2.5in]{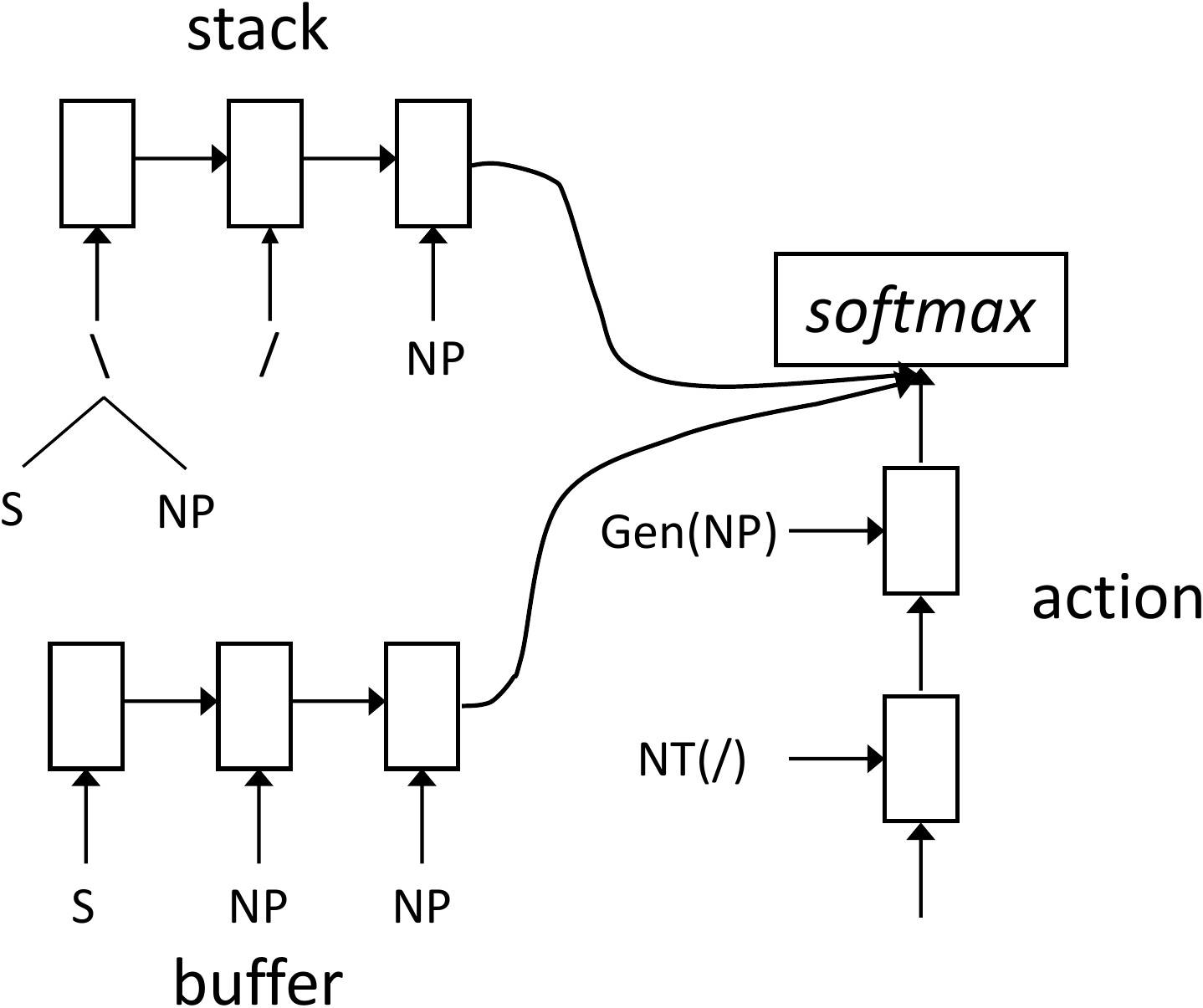}
    \caption{The architecture of our in-order system.}
    \label{fig:transition_system}
    \end{minipage}
    \begin{minipage}[t]{0.45\textwidth}
        \centering
        \includegraphics[width=2.5in]{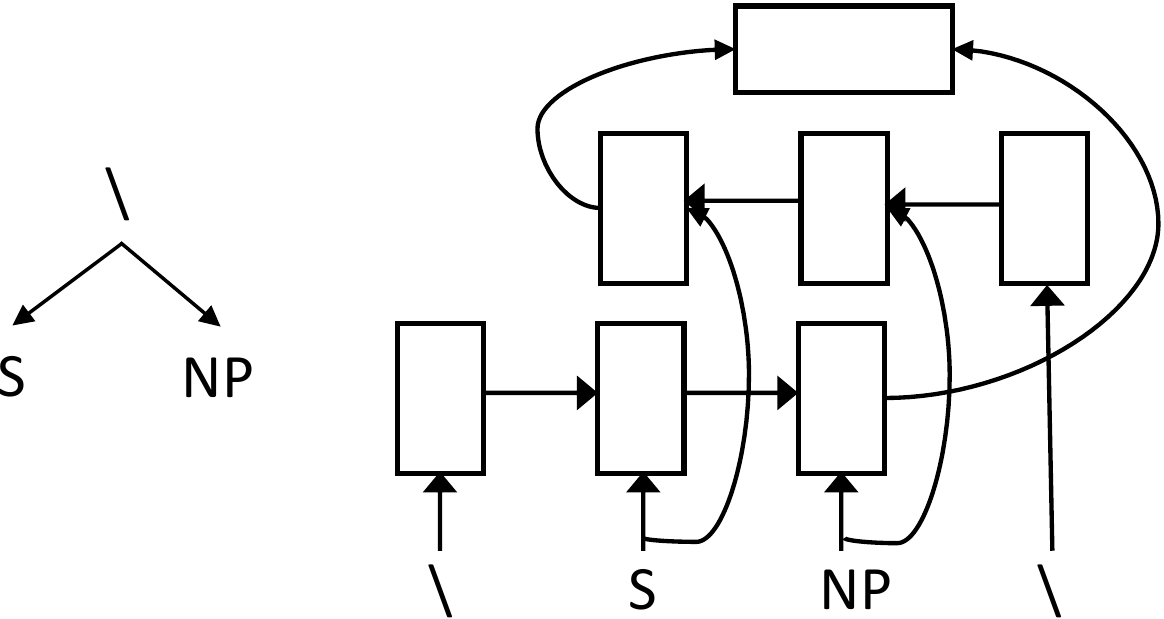}
        \caption{The composition function used in our model.}
        \label{fig:composition}
    \end{minipage}
\end{figure}

Following \citet{liu-zhang-2017-order}, we use LSTMs to obtain the representation. 
Specifically, a stack-LSTM is used to represent stack states, and vanilla LSTMs 
for output buffer and history actions. Besides, we employ a bidirectional LSTM to 
get the representation of the subtree. The model architecture is shown 
in Figure \ref{fig:transition_system}.

For a sentence word $x_i$, the probability of current action is defined as,
\begin{IEEEeqnarray*}{c}
    p(b_i^j = b|\mathbf{x)} = \underset{b \in \mathcal{T}_b}{\mathrm{Softmax}} ~
    \left(w ^\intercal \mathrm{Relu} [s_i^j; h_i] \right),
\end{IEEEeqnarray*}
where $s_i^j$ represents the information of the current state and 
$\mathcal{T}_b$ is the action set. The information of the current state
is computed as,
\begin{IEEEeqnarray*}{c}
    s_i^j = \left[ h_{\text{stack}}; ~ ~ h_{\text{out}}; ~ ~  h_{\text{his}} \right],
\end{IEEEeqnarray*}
where $h_{\text{stack}}$, $h_{\text{out}}$, $h_{\text{his}}$ represents the information of 
stack, output buffer and history actions. We use the output of the last hidden state of LSTMs.
Following \citet{liu-zhang-2017-order}, We employ a bidirectional LSTM as the composition function 
to represent subtree on stack, as shown in Figure \ref{fig:composition}. 
The representation of the subtree is computed as,
\begin{IEEEeqnarray*}{c}
    e_{root}^f = \mathrm{LSTM}(e_{\text{root}},e_1, e_2),\\
    e_{root}^b = \mathrm{LSTM}(e_{\text{root}},e_2, e_1) ,\\
    e_{root}^{'} = \mathrm{Relu} \left( w ^\intercal [ e_{root}^f; e_{root}^b] \right),\\
\end{IEEEeqnarray*}
where $e_1$, $e_2$ represent the embedding of left child and right child
and root node is actually the non-terminals. when a subtree is constructed, 
we push the subtree onto the stack with its representation $e_{root}^{'}$.

\section{Hyperparameters}
In word representations,
character embedding $c_i$ is calculated by CNN 
with $100$ filters and window size $3$. 
Pre-trained embeddings $\bar{w}_i$ are initialized with scaled $50$-dimensional
Turian embeddings \citep{DBLP:conf/acl/TurianRB10}.
The dimension of randomly initialized $w_i$ is $100$.

We also use BERT \cite{DBLP:conf/naacl/DevlinCLT19}
 (uncased, $24$ layers, $16$ attention
heads per layer and $1024$-dimensional hidden
vectors) and use the output of the last layer as the
pre-trained word embeddings. After linear transformation,
the BERT word embeddings are projected into $100$ dimensional vectors
and are concatenated with vectors $x_i$ as the final word representations.

The two-layer Bi-LSTM encoder has $400$ dimensional hidden vectors. 
For category generator, the dimension of LSTM hidden vectors is $250$, 
and the dimension of atomic tag embeddings is $30$.
For our transition model, the dimension of LSTM hidden vectors 
for stack, output buffer, action history and composition function
is $128$.

For the Italian CCGBank, We use the fastText\cite{bojanowski2017enriching} 
word embedding for pre-trained embedding\footnote{Download from 
\url{https://fasttext.cc/docs/en/crawl-vectors.html}}
and resize it's dimension to $50$ by linear transformation.
The dimension of randomly initialized $w_i$ is $50$ and 
the filters used in CNN is changed to $50$. 
The encoder has $200$ dimensional hidden vectors and 
the dimension of decoder LSTM hidden vectors is $100$.
Results are shown in Table \ref{tab:italian_results}.

We use Adam Trainer with 
learning rate $0.002$, $\beta_1 = 0.9$ and $\beta_2 = 0.9$,
batch size $200$(except transition model, batch size is $128$),
and set dropout rate $0.33$.
We use the java version\cite{clark2015java} of C\&C parser\cite{DBLP:journals/coling/ClarkC07}
with a beam threshold of $0.075$. 
The POS tags used in our models are automatically assigned by Stanford Tagger.
All experiments are conducted on 
a single 1.7GHz core and a single NVIDIA TITAN Xp GPU.


\begin{table}[t]
    \small
  \centering
  \begin{tabular}{lll}
  \toprule
  \textbf{Model} &\textbf{Dev} &\textbf{Test} \\
     \midrule
      CC  &63.33 &62.67 \\
      CG &  63.17 & 62.84\\
      CGNG2 & 64.55& 64.10$^*$ \\
      rerank & \textbf{64.67}$^*$ & \textbf{64.38}$^*$\\
    \bottomrule
  \end{tabular}
  \caption{Results on Italian CCGBank. 
  The results are averaged over 3 runs.
  $^*$ denotes the difference
  between one model and the CC model is significant.}
  \label{tab:italian_results}
  \end{table}


\bibliography{sup_aaai}